\documentclass[conference]{IEEEtran}
\IEEEoverridecommandlockouts

\usepackage{lipsum}  
\usepackage{caption}
\usepackage{subcaption}
\usepackage{cite}
\usepackage{amsmath,amssymb,amsfonts}
\usepackage{algorithmic}
\usepackage{graphicx}
\usepackage{subcaption} 
\usepackage{textcomp}
\usepackage{xcolor}
\usepackage[hidelinks]{hyperref}
\usepackage{orcidlink}

\def\BibTeX{{\rm B\kern-.05em{\sc i\kern-.025em b}\kern-.08em
    T\kern-.1667em\lower.7ex\hbox{E}\kern-.125emX}}
\begin{document}

\title{Navigating Time's Possibilities: Plausible Counterfactual Explanations for Multivariate Time-Series Forecast through Genetic Algorithms
}


\author{
    \IEEEauthorblockN{Gianlucca Zuin~\orcidlink{0000-0002-0429-3280} and Adriano Veloso~\orcidlink{0000-0002-9177-4954}}
    \IEEEauthorblockA{Universidade Federal de Minas Gerais, C.S. Dept. Belo Horizonte, Brazil.
    \IEEEauthorblockA{Kunumi. Belo Horizonte, Brazil.
    \\\{gianlucca, adriano\}@kunumi.com}
}}

\maketitle

\begin{abstract}

Counterfactual learning has become promising for understanding and modeling causality in complex and dynamic systems. This paper presents a novel method for counterfactual learning in the context of multivariate time series analysis and forecast. The primary objective is to uncover hidden causal relationships and identify potential interventions to achieve desired outcomes. The proposed methodology integrates genetic algorithms and rigorous causality tests to infer and validate counterfactual dependencies within temporal sequences. More specifically, we employ Granger causality to enhance the reliability of identified causal relationships, rigorously assessing their statistical significance. Then, genetic algorithms, in conjunction with quantile regression, are used to exploit these intricate causal relationships to project future scenarios. The synergy between genetic algorithms and causality tests ensures a thorough exploration of the temporal dynamics present in the data, revealing hidden dependencies and enabling the projection of outcomes under hypothetical interventions. We evaluate the performance of our algorithm on real-world data, showcasing its ability to handle complex causal relationships, revealing meaningful counterfactual insights, and allowing for the prediction of outcomes under hypothetical interventions.
\end{abstract}


\section{Introduction}
\emergencystretch 3em

Human fascination with alternative realities has captivated our imagination for centuries. From the pages of science fiction novels to the silver screen, the idea of exploring different possible worlds and divergent futures has always intrigued and entertained us. For instance, the notorious remark from Pascal, \textit{"Le nez de Cléopâtre si'l eût été plus court toute la face de la terre aurait changé"} (\textit{If Cleopatra's nose had been shorter, the whole face of the earth would have changed})~\cite{pascal1852pensees}.
 In works of fiction, authors have skillfully and imaginatively crafted narratives that transport us to parallel universes, offering glimpses into worlds where history unfolded differently, where characters make different choices, and where the course of events diverges from our own~\cite{prucher2007brave}.

While the idea of traversing parallel dimensions may remain in the realm of fiction, the desire to understand and explore alternative realities extends beyond storytelling. In the realm of data analysis and predictive modeling, researchers and practitioners seek to understand and explain the relationships of cause and effect within complex systems, such as time-series data~\cite{schlegel2019towards, theissler2022explainable}. 
Understanding the factors that influence the dynamics of time-series data is crucial in various domains, ranging from finance and economics~\cite{clements1998forecasting} to healthcare~\cite{zuin2022prediction}, 
and climate science~\cite{zuin2023extreme}. Time-series models provide valuable insights into the underlying patterns and trends within the data, enabling accurate predictions of future behavior. However, interpreting the complex relationships among variables in time-series data and uncovering the underlying mechanisms that drive specific outcomes remain challenging tasks~\cite{ghosh2020time}.


Among the various explanation approaches, counterfactuals have earned substantial attention due to their ability to provide insights that can aid the provision of interpretable models, making the decisions of complex systems intelligible to both developers and users.
Counterfactual explanations, in the context of time series, aim to address hypothetical questions about the outcomes that could have arisen if specific variables had taken different values or if certain events happened differently~\cite{byrne2019counterfactuals}. Often referred to as "What-If" scenarios, these techniques have emerged as an approach to enhance explicability~\cite{wexler2019if}. In this context, counterfactual modeling in time-series data faces significant challenges. Such systems contain various interdependent variables that contribute to outcomes collectively. Moreover, the temporal aspect of time-series data introduces lagged effects and non-linear dynamics.

We propose a method for generating counterfactual explanations in time-series data. 
In particular, we are concerned with multivariate counterfactual forecasts. That is, given a target endogenous variable and a hypothetical projected forecast, which paths should all the remaining exogenous variables follow to reach the said future. To generate future scenarios, we create auto-regressive models for each exogenous input variable, allowing us to forecast their values into the future. By incorporating these forecasts as inputs to a main model, we obtain a comprehensive forecast that projects the entire data into the future. However, in the context of counterfactual explanations in which a specific desired outcome is sought, we need to find future scenarios that approximate it within a certain error tolerance margin.

Thus, we also consider the range of plausible yet less probable forecasts for every input variable. This is a key premise that is often found lacking in related literature. To navigate the space of plausible future scenarios, we train multiple quantile regressors as independent models for each input variable, capturing their distributional information. We then employ a genetic algorithm~\cite{konak2006multi}, which evolves a population of potential solutions. This combination of methods enables the search inside the large space of possible futures, projecting scenarios that are coherent with past historical data and that satisfy desired multivariate counterfactual forecast and explanation constraints. Our study was conducted in partnership with M. Dias Branco, a leading food company in Latin America. Our study employed real-world data from their operations, specifically concerning vacuum breaks in the deodorizer process. Briefly, vacuum breaks can critically disrupt the deodorization process, causing food quality issues and operational inefficiencies. To mitigate and recover from such events, we integrated the forecast models with real-time monitoring systems, enabling detection and response.




\section{Related Work}



Counterfactual explanations offer numerous advantages but also present challenges \cite{sokol2019counterfactual, stepin2021survey}. They are easily understandable to humans, enabling users to grasp the changes needed to attain a desired outcome. By introducing new information to existing facts, counterfactual explanations are informative and foster innovative problem-solving. Furthermore, they can shed light on the decision-making process, empowering organizations to ensure fairness and compliance with regulations. However, the existence of multiple potential explanations for a given outcome, also known as the "Rashomon effect"~\cite{breiman2001statistical, fisher2019rashomon, zuin2021predicting, zuin2023ensemble, ctd}, may turn the selection of appropriate ones challenging. Determining which features are actionable and should be modified can be subjective and contingent upon the context.

\cite{molnar2020interpretable} surveys the counterfactual literature and highlights some key methods for counterfactual explanations. \cite{wachter2017counterfactual} proposes a method for generating counterfactual explanations by random sampling and minimizing a loss function. The loss consists of two terms: the quadratic distance between the predicted outcome of a counterfactual instance and the desired outcome, and a distance measure between the instance to be explained and the counterfactual. The optimization makes perturbations to the input data by searching the complete input space until a suitable and satisfactory counterfactual is found. However, this sort of approach often is associated with extrapolation and may lead to implausible explanations that lie outside of the data domain \cite{yang2020generating}. Another approach is to restrain the sampling space to plausible data, as proposed by \cite{mothilal2020dice}. We can restrict ourselves to looking only to past instances, and search for neighbors in the multi-dimensional feature space that satisfies the counterfactual example. However, these methods may not be suitable for time series data.

In the context of multivariate time series, the application of the method proposed by \cite{wachter2017counterfactual} raises concerns. Treating each time state of the variables as distinct features during the optimization of the loss function can disrupt the time dependence between states, resulting in implausible counterfactuals. Consequently, this approach is not suitable for addressing such problems. However, the other option, the method presented by \cite{mothilal2020dice}, is also unsuitable. To apply this method, the time series would need to be flattened through vectorization. However, this process significantly increases the dimensionality of the data, making it challenging to find past instances that are both suitable and closely aligned with the current scenario. 

Many other works tackle the challenge of counterfactual explanations for multivariate time series, however, all lack some key components that distinguish our work from theirs. For instance, \cite{lang2023generating} uses GANS to create counterfactual scenarios for an entire time series representing the dataset as a heatmap, but only tackles the classification problems. The same can be said of \cite{bahri2022temporal} and \cite{ates2021counterfactual} who likewise focus on classification. To the author's knowledge, few methods focus on regression forecasts, and even fewer project the target continuous variable onto the future alongside the input variables. As such, we were not able to find any strong baseline in the literature.


It is also important to mention other counterfactual approaches that rely on similar methods to ours, albeit not necessarily applicable to time series. \cite{sharma2020geneticcounterfactual} proposes CERTIFAI, a model-agnostic and data-agnostic approach for generating counterfactual explanations that use a population of candidate counterfactuals; MOC (Multi-Objective Counterfactuals), presented by \cite{back2020parallel} which also leverages a genetic algorithm but formulates the counterfactual search as a multi-objective optimization problem; or thee GEnetic COunterfactual explainer (GECO) introduced by \cite{schleich2021geco} that differentiates itself by automatically accounting for plausibility and accountability during mutation and crossover operations. 
However, some key differences should be mentioned. While CERTIFAI, MOC, and GECO are designed for tabular data, our approach is specifically tailored for time series data. We incorporate the Granger causality test to reveal causal relationships among variables, crucial for capturing temporal dependencies. Additionally, we employ quantile regressions to capture uncertainty and variability in time series, providing a comprehensive view of potential future trajectories. 

\section{Proposed Approach}

Counterfactuals, in the context of time series, refer to hypothetical scenarios where the observed sequence of events is altered in some way, allowing for an analysis of alternative outcomes. Formally, let $X=\{X_1,X_2,...,X_T\}$ represent a multivariate time series consisting of $T$ observations, where $X_t$ denotes the value of the time series at time $t$ and where each $X_t$ can be further divided into $x_{t1}, x_{t2}, ... x_{tV} $ representing each of the $V$ variables at time $t$.

A counterfactual scenario $X'=\{X_1',X_2',...,X_N'\}$ with $N \leq T$ is constructed by manipulating the original time series $X$ by a predefined intervention. Thus, let $A$ denote the set of feasible interventions that can be applied to the time series $X$, and $P(A)$ denote the power set of $A$, representing all possible combinations of interventions. For a given counterfactual scenario $X'$, the counterfactual outcome $Y'=\{Y_1',Y_2',...,Y_N'\}$ is determined by a causal model $g:X\times A\rightarrow Y$. The causal model $g$ captures the causal relationships between the time series data $X$ and the outcome $Y$ under different interventions. Mathematically, it can be represented as:

\begin{equation}
Y' = g(X', A) 
\end{equation}

The main challenge is two-fold. The first is the selection of a subset $A \subseteq P(A)$ such that only realistic and feasible interventions are contemplated. We consider an intervention feasible if it maintains consistency with past values of $X$, ensuring that the distribution of $X'$ follows the one observed in $X$. The second challenge consists of finding a group of interventions $a \in A$ that satisfies some constraint in regards to some unknown $Y'$. For example, we can establish a margin $\epsilon$ and some set of target values $\mathcal{Y}$, and posit the constraint $\mathcal{Y} - \epsilon \leq Y' \leq \mathcal{Y} + \epsilon$. Now, let $\hat{f}(X)$ denote a forecasting model that predicts future target values of the time series variables based on the historical data $X$. Then, a counterfactual forecast $\hat{f}(X')$ represents the predicted future values under the altered scenario $X'$, which enables the evaluation of $\hat{Y'}$ that can be used to select an intervention $a$ that approximates the desired constraints.

To restrain the search for plausible future scenarios and obtain a suitable $A$, we propose training many quantile regressors as independent forecasting models for each input variable. This is achieved by using past historical data and optimizing the pinball loss \cite{steinwart2011estimating}. Instead of predicting a single-point estimate, the goal is to estimate a range and the loss evaluates how well the predicted quantiles match the actual quantiles of the target variable. The pinball loss is used to evaluate how well the predicted quantiles match the actual quantiles of the target variable, defined as follows for a given quantile level $\tau$ (ranging between 0 and 1) and a target variable $x$:

\begin{equation}
\mathcal{L}(\tau, x) = \max(\tau \cdot (x - x_{\text{pred}}), (\tau - 1) 
\end{equation}
\noindent where $x_{\text{pred}}$ represents the predicted value for the target variable $x$. The pinball loss penalizes underestimations when $x$ is above $x_{\text{pred}}$ and overestimations when $x$ is below $x_{\text{pred}}$, with the magnitude of the penalty increasing as $\tau$ moves away from 0.5. 

By minimizing the pinball loss, quantile models can learn how to capture the full distributional information. 
Once we have trained these regressors, we can leverage them to search the input space. The intervention policy $a$, tied to a future scenario projection, can be described as a $t$-shaped vector where $t = V \times N$. Each index corresponds to the probability (quantile) used to obtain the forecast of each $x_{nv} \in X'$ for each time-state $n \in N$ and variable $v \in V$.

We start by creating a one-step-ahead auto-regressive model, denoted as $\hat{f}^{x}(X, \tau)$, for each input variable $x \in X$. When we apply an intervention policy $a$ and forecast each variable recursively $N$ steps into the future, we generate a set of predicted values $\hat{X}'={\hat{X}_1, \hat{X}_2, ..., \hat{X}_{N}}$. This set represents our projections for each variable at different time steps, and which can be used to estimate the target variable in the future. 
To enhance the accuracy of our projected values $\hat{X}'$, we employ the Granger causality test to learn the causal relationships among pairs of input features. The main goal of our Granger test is to reduce the Rashomon space of feasible forecasts.

The Rashomon space refers to the set of all models that explain the data equally well in terms of predictive accuracy but may lead to different decisions or interpretations \cite{zuin2023ensemble}. In practical scenarios, this space can be quite large, making it difficult to derive actionable insights from the models due to variability in outcomes. By reducing the Rashomon space, we aim to narrow down the range of models to those that not only perform well but also provide stable, interpretable, and consistent predictions. By leveraging the Granger causality test, we focus only on causally related pairs of variables.

Our Granger causality test involves comparing the performance of two auto-regressive models: one with the addition of an extra evaluated feature and one without it. For each input variable, we train an auto-regressive model independently using past historical data. We then compare the performance of the auto-regressive model with the inclusion of the extra-evaluated feature. 

The null hypothesis assumes that the evaluated feature does not Granger-cause the target variable, and the p-value resulting from the t-test represents the probability of obtaining the observed improvement in prediction accuracy by chance. A p-value lower than the chosen significance level ($0.05$) indicates evidence against the null hypothesis and suggests the presence of Granger causality.
If the p-value from the t-test is lower than $0.05$, we assume Granger causality. We describe a future scenario as a $t$-shaped vector, where $t = |\mathcal{X}| \times N$. Each index corresponds to the probability (quantile) used to obtain the forecast of $x \in X$ for each time state $n \in N$. The auto-regressive model with the addition of the evaluated feature can be represented by:

\begin{equation}
\hat{y}i = \alpha + \sum{j=1}^{p} \beta_j y_{i-j} + \sum_{j=1}^{p} \gamma_j x_{j,i-j} + \varepsilon_i
\end{equation}

\noindent where $\hat{y}i$ is the predicted value of the target variable at time $i$, $y{i-j}$ represents the lagged values of the target variable up to order $p$, $x_{j,i-j}$ denotes the lagged values of the evaluated feature $x_j$ up to order $p$, $\alpha$ is the intercept term, $\beta_j$ and $\gamma_j$ are the coefficients for the lagged values, and $\varepsilon_i$ is the error term. Similarly, the auto-regressive model without the evaluated feature can be represented by:

\begin{equation}
\hat{y}^{*}i = \alpha + \sum{j=1}^{p} \beta_j y_{i-j} + \varepsilon_i
\end{equation}

By comparing the performance of these two models using a t-test, we can determine if the evaluated feature has a significant impact on the target variable.

The proposed counterfactual algorithm aims to find a satisfiable projection. That is, an intervention that leads to a projection that approximates the desired $\mathcal{Y}$ under a small error margin $\epsilon$. We employ a genetic algorithm~\cite{michalak2021evolutionary} to search for the quantile sequence associated with each input variable. The algorithm mimics the process of natural selection to generate high-quality solutions to optimization and search problems. It iteratively evolves a population of candidate solutions by selecting, combining, and mutating them. We employ an adaptation of the fitness function proposed by \cite{back2020parallel} as a three-objective for counterfactual explanations. 
The first objective, $o_1$, minimizes the distance between the predicted counterfactual $\hat{f}({X}')$ and the desired $\mathcal{Y}_{\text{goal}}$:

\begin{equation}
o_1(\hat{f}({X}'), \mathcal{Y}_{\text{goal}}) = ||\hat{f}({X}') - \mathcal{Y}_{\text{goal}}||
\end{equation}

The second objective, $o_2$, measures the similarity between the counterfactual ${X}'$ and the original ${X}$:

\begin{equation}
o_2({X}, {X}') = 1 - \frac{{{X} \cdot {X}'}}{{||{X}|| \, ||{X}'||}}
\end{equation}

The third objective, $o_3$, evaluates the likelihood of ${X}'$:

\begin{equation}
o_3({X}', {X}_{\text{obs}}) = -\log P({X}'|{X}_{\text{obs}})
\end{equation}

We hypothesize that the counterfactual method described could allow for a better understanding of the projections of a time series model and the variables that drive future scenarios. By performing the Granger causality test between all variable pairs, we identify the features that impact one another. Incorporating these features, we create auto-regressive models to generate future scenarios. To ensure the plausibility of these scenarios, we train many quantile regressors and search the input space for inputs that satisfy specific outcome constraints. This combined approach enhances the predictive capabilities of our models, enabling the generation of plausible future scenarios aligned with desired outcomes.
\section{Experiments}

To evaluate our proposed method for generating counterfactual explanations in time series data, we conducted experiments using real-world data of M. Dias Branco, a food company holding a third of the Brazilian market-share in the biscuits and pasta fields~\cite{mdiasmarketshare}. We focus on their Alkaline Closed Loop (ACL) vacuum system. This is a specialized setup used in industries such as vacuum distillation, chemical processing, and other processes where maintaining specific vacuum conditions is crucial. 

\begin{figure}[!htb]
\centering
\includegraphics[width=0.99\linewidth]{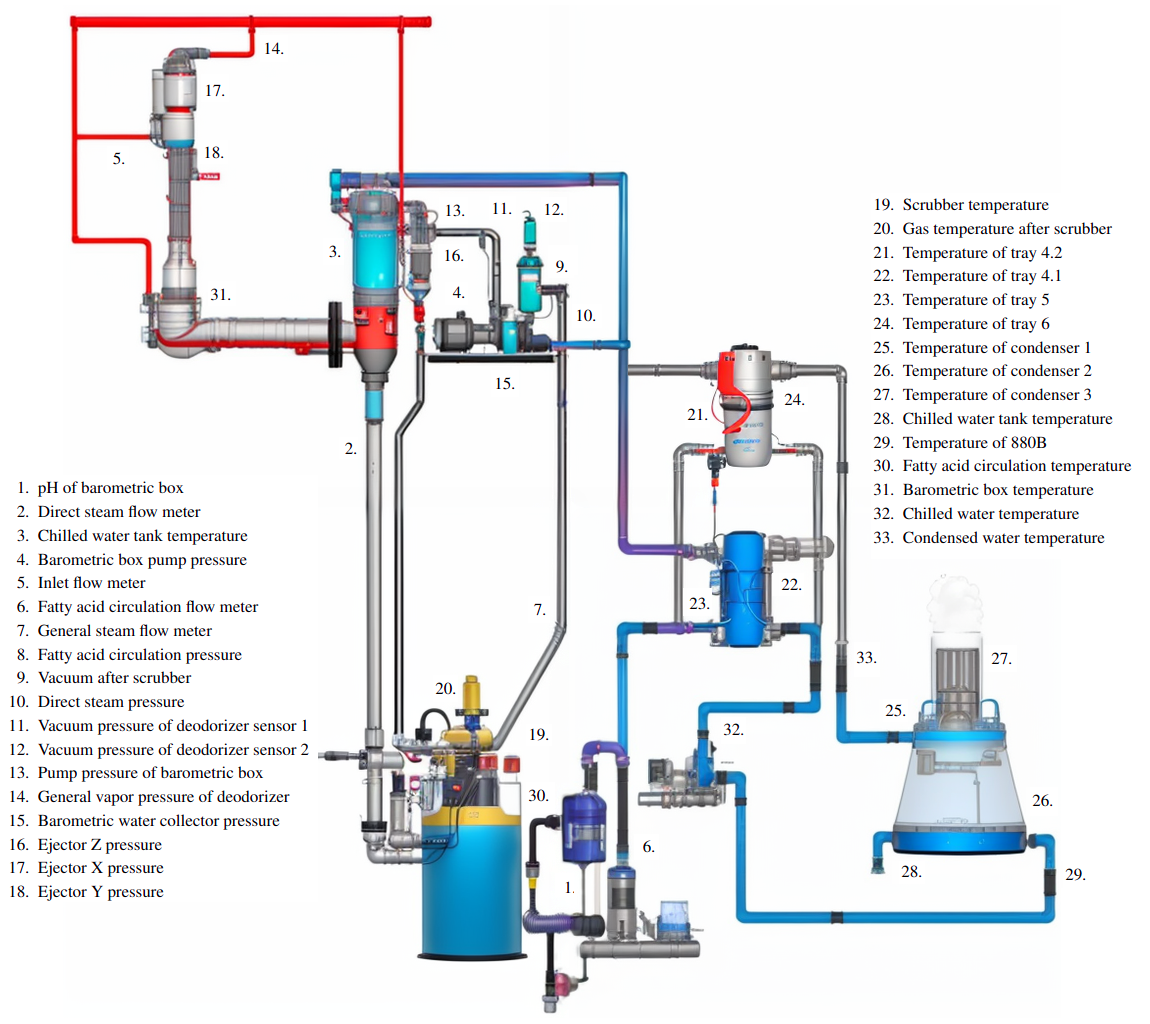}
\caption{Alkaline Closed Loop vacuum system.}
\label{fig:acl}
\end{figure}

Figure~\ref{fig:acl} shows how the main variables of the ACL system. In essence, it employs a closed-loop circuit to transfer heat from a vacuum chamber's equipment to a heat exchanger, where the heat is then transferred to cooling tower water for dissipation through evaporative cooling. The alkaline treatment of the cooling water is essential to prevent corrosion and ensure the stability of the vacuum conditions required for the specific industrial process. In M. Dias Branco, this step controls food odors by keeping the vacuum constant.

We aim to predict the pressure value $T$ timesteps into the future and find a set of input feature paths that approximate the queried counterfactual. To achieve this, we require the values of all input features for each of the preceding $T-1$ timesteps. However, there are many possible paths that each feature might take. 
This leads to a vector of size $T * V$ which is the input to our genetic algorithm. Our fitness function aims to find an individual (a set of quantiles) that produces a forecast that matches the queried scenario while minimizing divergence from the present. To quantify these objectives, we calculate their absolute differences. Once found, we can verify which variables are actionable and perform interventions.

\vspace{0.05in}
\noindent \textbf{Data and ML Models:} Our dataset consists of the 33 time-dependent variables shown in Figure~\ref{fig:acl}. Each variable has its values recorded at 3-second intervals over several months. To ensure the validity and statistical significance of our results, we performed a 1-month walk-forward cross-validation, training our model in one month and validating performance on the following one. We compared the performance of several machine learning and statistical models. 
To assess their predictive capabilities, we employed the Mean Squared Error (MSE), Mean Absolute Error (MAE), and the Coefficient of determination ($R^2$) metrics. We also provided the relevant training and inference times for each approach. The results of our model comparison are summarized in Table \ref{tab:table}. Each model was trained to predict the target variable 30 seconds into the future on out-of-sample data.


\setlength{\tabcolsep}{0.53em}
\begin{table}[!t]
\centering
\caption{Model comparison and predictive performance.}
\label{tab:table}
\small{
\centering
\begin{tabular}{lccccc}
\\\hline
\multicolumn{1}{l|}{Model}             & MAE & MSE & \multicolumn{1}{c|}{$\text{R}^2$} & \begin{tabular}[c]{@{}c@{}}Training\\ Time (s)\end{tabular} & \begin{tabular}[c]{@{}c@{}}Inference\\ Time (s)\end{tabular} \\ \hline
\multicolumn{1}{l|}{lightGBM}  &   \textbf{0.889} &  \textbf{2.210} &  \multicolumn{1}{l|}{\textbf{0.505}} & 37.81 &   2.55       \\
\multicolumn{1}{l|}{Linear Reg.} &  1.055   &  2.375    & \multicolumn{1}{l|}{0.401}                     &    \textbf{6.88}                                                     & \textbf{2.03}                                                          \\
\multicolumn{1}{l|}{Elastic Net} &  1.082   &  \textbf{2.179}    & \multicolumn{1}{l|}{0.393}                     &    9.52                                                     & \textbf{2.17}                                                          \\
\multicolumn{1}{l|}{SVR} &   1.393  &   5.398   & \multicolumn{1}{l|}{0.183}                     &  432.65                                                       &    529.39                                                      \\
\multicolumn{1}{l|}{LSTM}              & 1.288     &  3.432    & \multicolumn{1}{l|}{0.074}                     &    1405.32                                                     &   7.76                                                       \\
\multicolumn{1}{l|}{LSTM+Bias}  &  1.259    &  3.563   &  \multicolumn{1}{l|}{0.108} &  544.65      &     7.26s                                                                                                               \\
\multicolumn{1}{l|}{LSTM+CNN}  &  1.616    &  4.788   &  \multicolumn{1}{l|}{0.015} & 305.01      &                16.94                                                                                                               \\
\hline
\end{tabular}
}
\end{table}

\begin{figure*}
\centering
\begin{subfigure}{0.49\linewidth}
  \centering
  \includegraphics[width=0.95\linewidth, height=8.5cm]{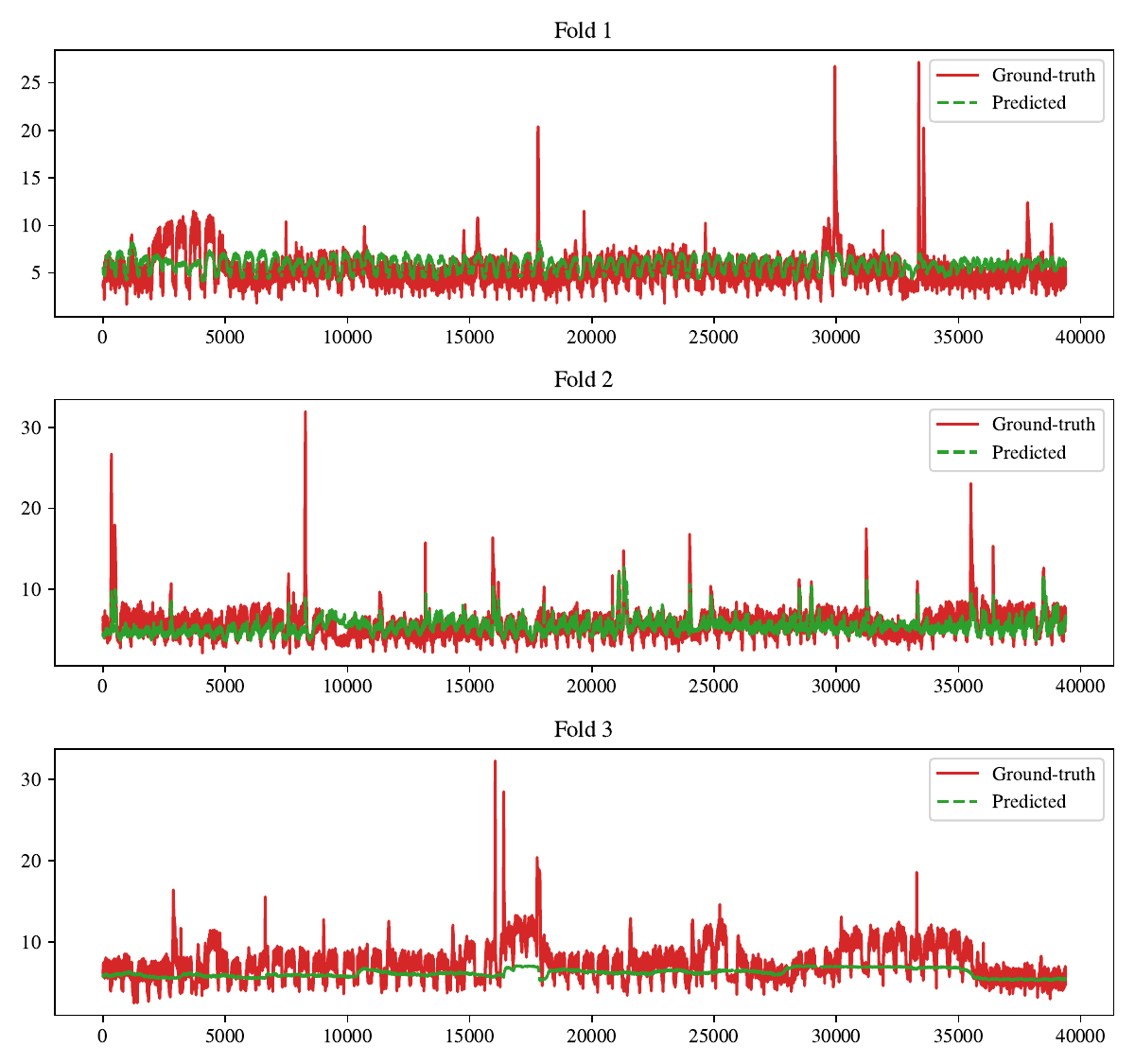}
  \caption{LSTM.}
  \label{fig:subfigure1}
\end{subfigure}%
\hspace{0.01\linewidth}
\begin{subfigure}{0.49\linewidth}
  \centering
 \includegraphics[width=0.95\linewidth, height=8.5cm]{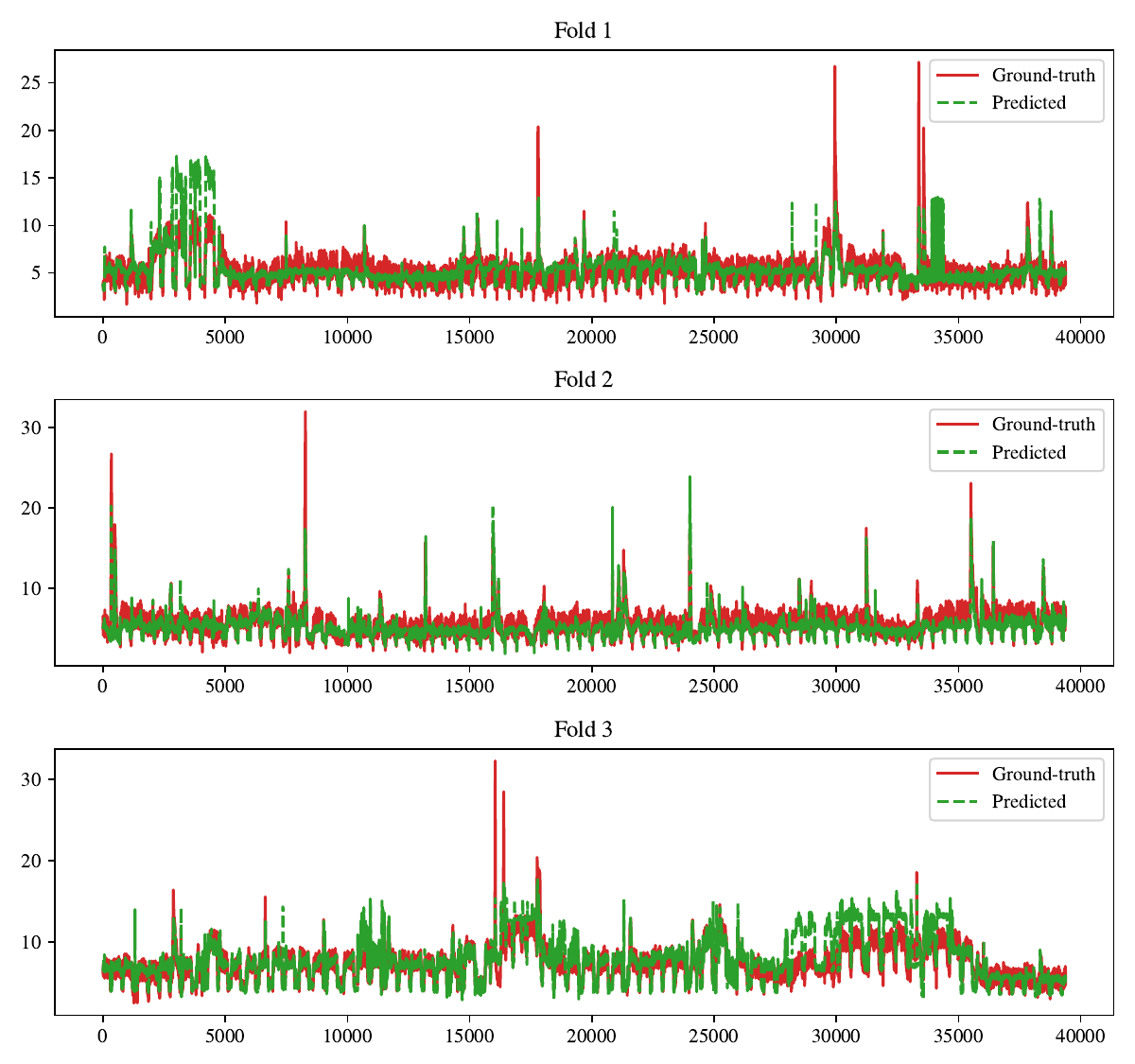}
  \caption{lightGBM.}
  \label{fig:subfigure2}
\end{subfigure}
\vspace*{-0.2cm}
\caption{Prediction and ground-truth values during cross-validation for a subset of the data.}
\label{fig:cross_validation_predictions}
\end{figure*}

We also compared the performance of ARIMA~\cite{siami2018comparison} and the performant lightGBM separately. We used the data up to the latest point to predict future steps. In this evaluation, based on the latter 83 hours of our dataset (10,000 data points), lightGBM performed better than ARIMA. It achieved a MAPE of $3.8\%$, an $R^2$ value of $0.916$, and a MSE of $1.602$. On the other hand, the ARIMA model yielded a MAPE of $5.20\%$, an $R^2$ of $0.914$, and an MSE of $1.557$. We decided to select the lightGBM as our chosen learning algorithm due to its lower MAPE and explicability capabilities, albeit slower.

To gain further insights into the models' performance, we visualized the predictions made by the LSTM and lightGBM models on the most recent folds during cross-validation, as depicted in Figure \ref{fig:cross_validation_predictions}. The plot showcases the predicted values (represented by a dashed line) and the corresponding ground-truth values (solid line). From the plot, we observe that the predictions made by the lightGBM model more closely follow the actual ground-truth values, exhibiting a strong alignment with the underlying data patterns. On the other hand, the LSTM model's predictions appear to not be as fine-grained, following more closely the mean of the period and not properly capturing the minor variations in the target variable, thus resulting in slightly lower predictive performance. The cross-validation analysis also confirms the robustness of the model, as it consistently outperforms the alternative model across different folds, further supporting its suitability for accurate time series prediction tasks.

\vspace{0.05in}
\noindent \textbf{Causal relationship analysis:} We conducted a comprehensive Granger causality analysis on all variable pairs. The Granger causality test allowed the determination of which past values of a time series contain valuable information for predicting another. We examined the resulting heatmap and marked cells corresponding to variable pairs with p-values less than or equal to 0.05 after a pair-wise t-test.


\begin{figure}
\centering
\includegraphics[width=\linewidth, height=8cm]{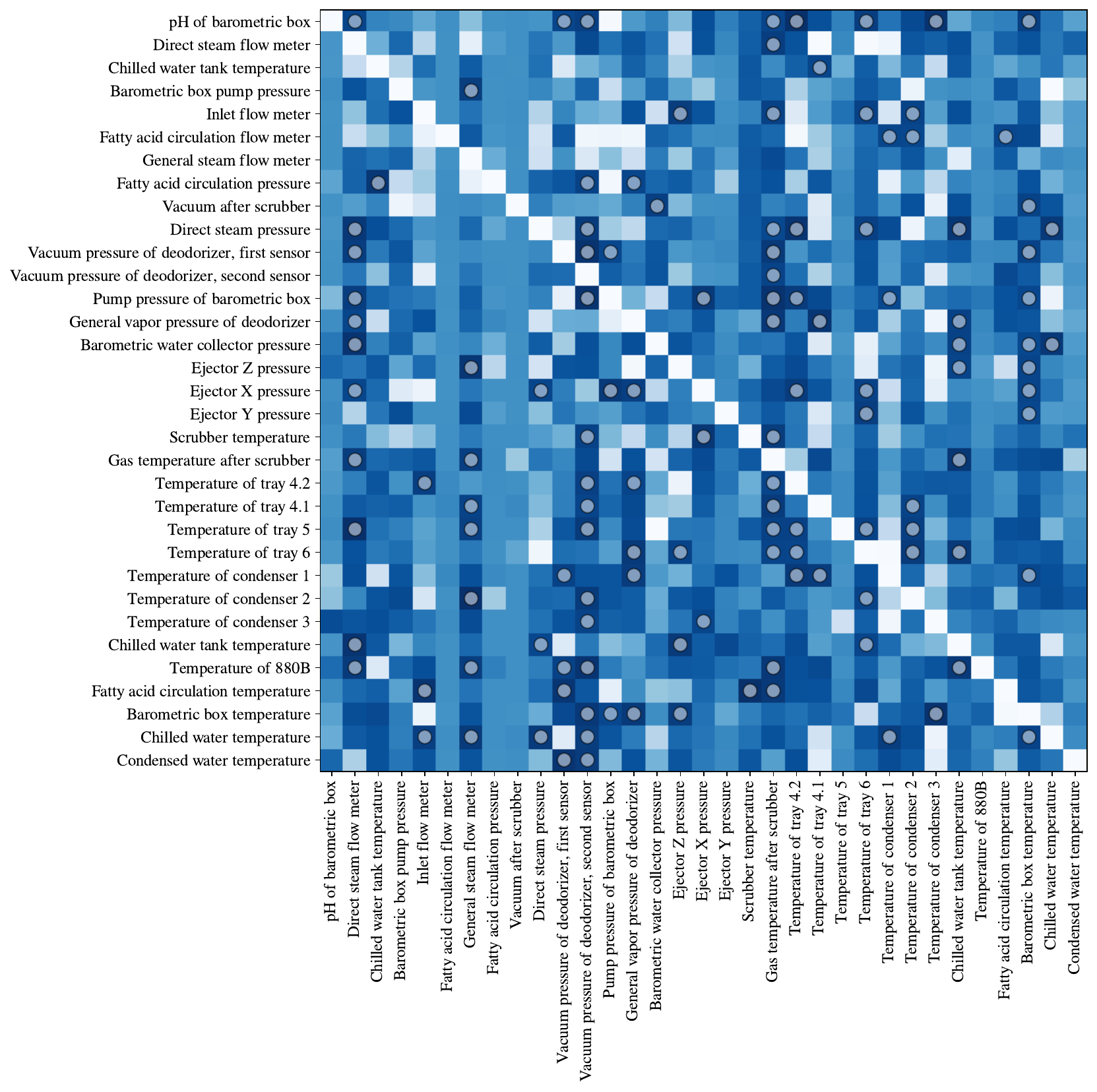}
\vspace*{-0.1cm}
\caption{Granger Causality Heatmap. Darker cells are associated with a stronger relationship with statistically significant pairs being marked.}
\label{fig:granger_heatmap}
\vspace*{-0.3cm}
\end{figure}

Figure \ref{fig:granger_heatmap} showcases the results of our Granger causality analysis in the form of a heatmap. The visualization highlights the interconnectedness of variables, represented by the varying intensities of the heatmap cells. The darker cells correspond to pairs of variables with statistically significant causal relationships, while lighter cells indicate weaker or insignificant connections. By examining the heatmap, we can discern clusters of variables that exert strong causal influence over each other, forming interconnected patterns that shape the behavior of the entire explored phenomena.

\begin{figure}[!h]
\centering
\includegraphics[width=0.95\linewidth]{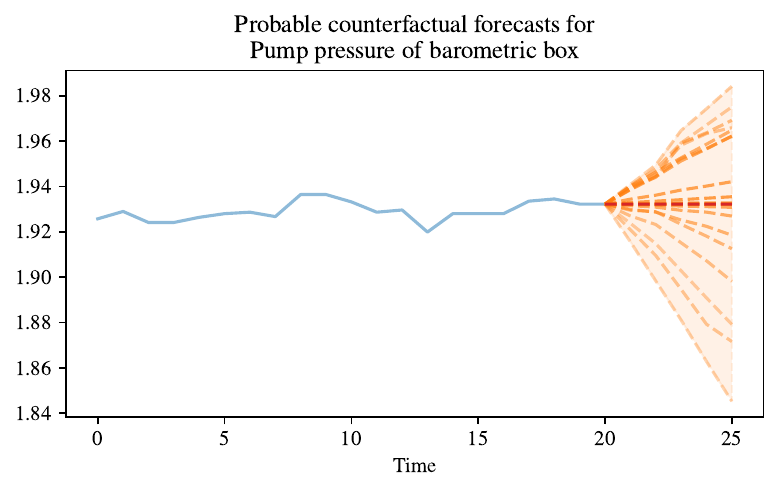}
\vspace*{-0.15cm}
\caption{Quantile forecasts for an actionable variables and some possible future pathways. Solid lines represent the historical values, while dashed lines illustrate the possible futures. The red dashed line highlights the most likely pathway. 
Specifically, values in the range [1.84, 1.98] are plausible for 15 seconds (5 steps) into the future. 
}
\label{fig:quantile_forecasts}
\end{figure}

\vspace{0.05in}
\noindent \textbf{Quantile regression for plausible scenarios:} To search for plausible future scenarios, we trained multiple quantile regressors for each input variable using historical data. These quantile regressors estimate the conditional quantiles of each target variable given the input features, capturing the distributional information. Quantile regression offers a powerful framework for quantifying uncertainty and exploring various possible future trajectories. By estimating conditional quantiles, we obtain a range of plausible values as in Figure \ref{fig:quantile_forecasts}.

\vspace{0.05in}
\noindent \textbf{Genetic Algorithm search:} To find future scenarios that satisfy specific outcome constraints, we employed a genetic algorithm with random immigrants. The genetic algorithm was configured with a population size of $200$, a mutation probability of $0.25$, a crossover probability of $0.75$, a tournament size of $3$, a constant immigration rate of $10\%$, and $100$ generations. The tolerance margin was set to $5\%$ for an early stop, and the time horizon for searching future scenarios was $30$ seconds. We define each individual as a vector, in which each position represents the quantile for each variable and for each time step into the future.

\begin{figure}
    \centering\begin{subfigure}[b]{0.95\linewidth}
        \centering
        \includegraphics[width=0.9\linewidth, height=5cm]{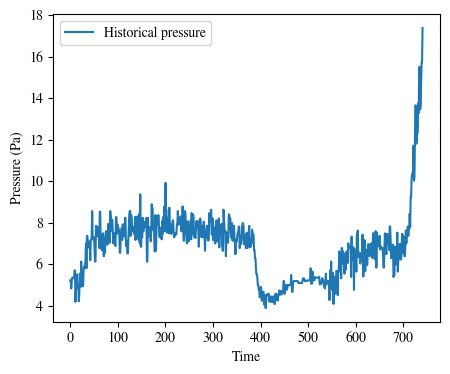}
        \caption{Time-series near a vacuum break.}
        \label{fig:vacuumbreak}
    \end{subfigure}
    
    \hfill
    
    \begin{subfigure}[b]{0.95\linewidth}
        \centering
        \includegraphics[width=0.9\linewidth, height=5cm]{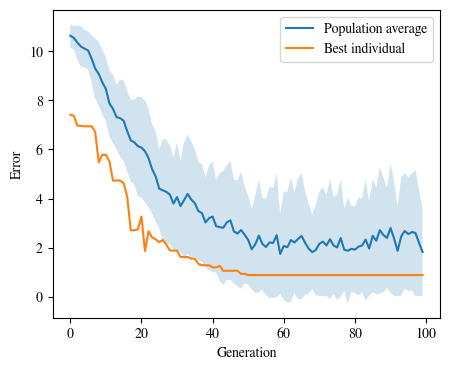}
        \caption{Fitness evolution of the Genetic Algorithm.}
        \label{fig:fitness_evolutiona}
    \end{subfigure}
    \caption{Population fitness during the GA execution projecting a plausible return from a near vacuum break. While searching for a near future where the pressure level reaches 5.0 Pa. The solution found lies in the range between 4.0 Pa and 6.0 Pa, satisfying the constraint of recovering from a vacuum break.}
    \label{fig:fitness_evolution}
\end{figure}

We conducted experiments to set the population size, finding that populations with fewer than 50 individuals had difficulty converging to good solutions. Further, larger populations led to quicker convergence (a population of 100 individuals converges near gen. 80 while one with 500 converges near gen. 20), but no significant gains were observed beyond 500 individuals. The chosen population and generations parameter pair were selected for visualization purposes; in reality, convergence occurs much earlier as seen in Figure \ref{fig:fitness_evolution}. 
The remaining parameters were chosen by drawing from literature.


\begin{figure*}
\centering
\includegraphics[width=0.93\linewidth]{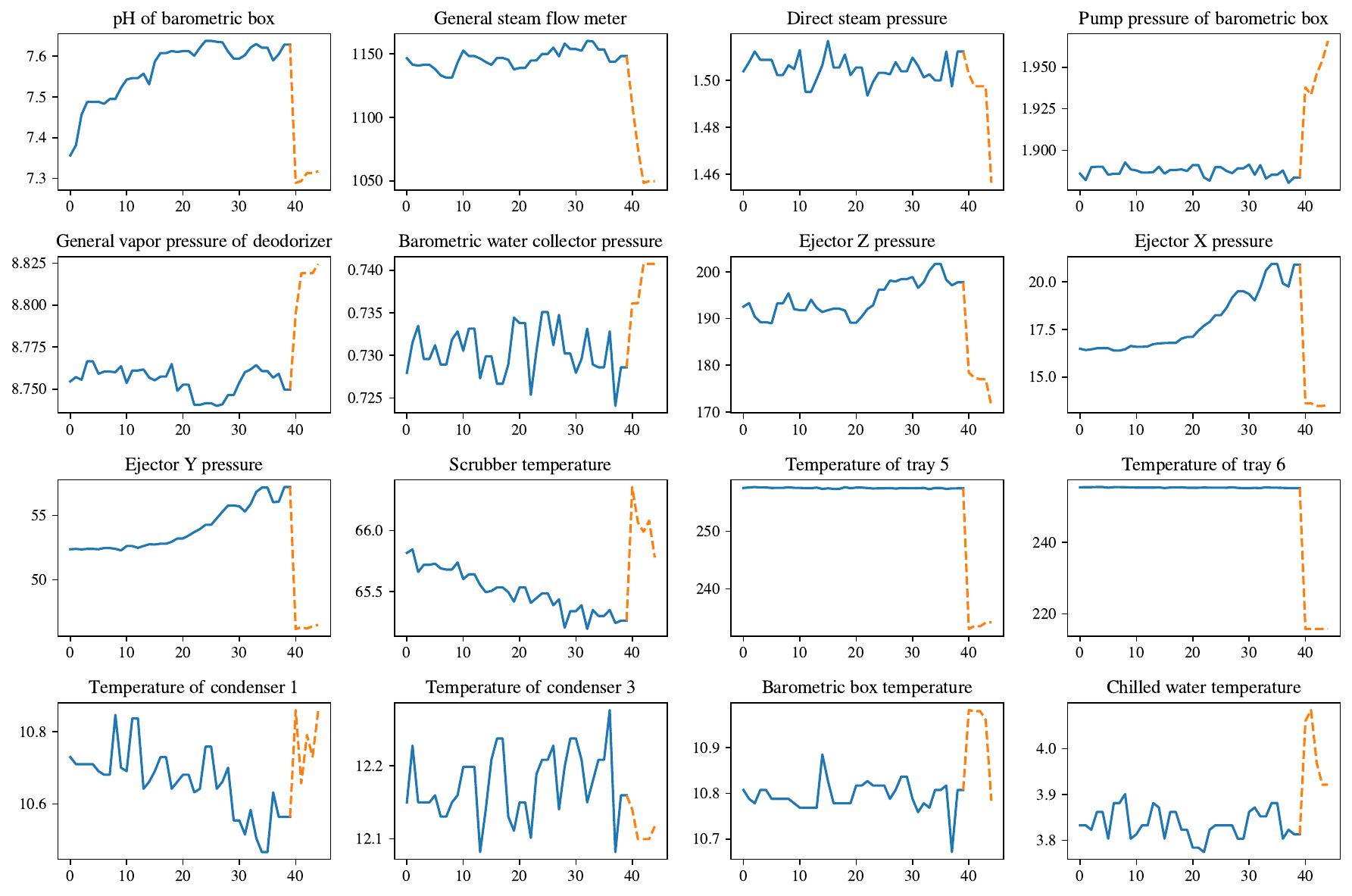}
\caption{Forecast of a future that enables the system to recover from a near vacuum break and satisfies the desired counterfactual explanation. Scenario likelihood: 42.9\%. Projected Pressure: 5.87 Pa.}
\label{fig:counterfactual_explanation}
\end{figure*}

The genetic algorithm explored the space of possible solutions and identified scenarios that satisfied the desired outcome within the tolerance margin. By iteratively refining solutions over generations, it was able to converge to counterfactual explanations that approximate the desired outcome. Figure \ref{fig:vacuumbreak} depicts a situation where the pressure on the ACL system is unusually high, which will lead to a vacuum break and food odor contamination. We utilize a genetic algorithm (GA) with a specific objective: achieving a future pressure value of 5.0 PA. The Y-axis in Figure \ref{fig:fitness_evolutiona} represents the error, showing the absolute difference between the forecasted values generated by individuals in each iteration and the desired target of 5.0 Pa. The resulting counterfactual explanation, shown in Figure \ref{fig:counterfactual_explanation}, provides a detailed insight into the key variables influencing the desired future state. For the given example, these insights could guide decision-makers in preventing a vacuum break and restoring system stability. Since we evolve an entire population, specialists can look in all of them for an actionable one, and Figure \ref{fig:counterfactual_explanation} illustrates such an example. We compute the scenario likelihood by the joint probability of the projection for each actionable variable. Specifically, we evaluate the distance from the 0.5 quantile in each time-step.


\section{Discussion}

Our experiments demonstrate the efficacy of the proposed approach for generating counterfactual explanations in time series data. By leveraging a combination of Granger causality analysis, quantile regression, and genetic algorithms, we successfully searched the solution space of possible future scenarios.
The Granger causality analysis unveiled the causal relationships among time series variables, providing an understanding of their interconnectedness. This was one of the key points to enable the plausibility of future scenarios. 

The genetic algorithm efficiently explored the space of possible solutions. The resulting counterfactual explanations offered a comprehensive view of the key variables that influence specific future trajectories, empowering decision-makers with actionable insights. It also enables simulations that explore possible preventative safety measures. 
When presented with the developed tool, the experts of M. Dias Branco were able to simulate a large amount of scenarios and with great enthusiasm, and were able to learn new patterns about their system and develop new countermeasures to prevent vacuum breaks. We conducted experiments repeating our entire pipeline 30 times, always retaining the population of 200 solutions and 100 generations. In over $90\%$ of our experiments (28/30), we were able to find a projected scenario that satisfied our constraints and could be used by the experts of M. Dias Branco, showcasing the robustness of the devised technique.

The most time-consuming steps in our approach are training the models and conducting the Granger tests. Inference during the GA is not as computationally intensive. In terms of time complexity, it is approximately $O(tdTVPG)$. Here, $O(td)$ denotes the complexity of making inferences with a GBM of $t$ trees with depth $d$. $T$ is the number of forecasted timesteps into the future, $V$ is the number of variables, $P$ is the population size and $G$ is the number of generations.

Our experiments also highlighted the role of model selection, as lightGBM consistently outperformed the alternatives. However, our observations also revealed a common challenge: a difficulty in extrapolation. Specifically, the models tended to undervalue points associated with extremes of the distribution. As we increased the amount of training data about these distribution extremes, the models demonstrated improved learning capabilities, evident when comparing the predictions between folds 1, 2, 3, and 4, as illustrated in Figure \ref{fig:cross_validation_predictions}.
Folds 1 and 3, which contained higher and more numerous peak values of the target variable, led to more accurate predictions in folds 2 and 4. On the other hand, predictions in fold 1 highlighted the models' limited capability in capturing these extreme values when the training data is not as representative of them.
\section{Conclusion}

In this work, we proposed a novel method for discovering counterfactual explanations for time series data, primarily leveraging genetic algorithms, quantile regressions, and Granger causality. Our approach offers a framework for navigating the data space containing multiple potential forecast projections while searching for counterfactual explanations. When presented to experts from M. Dias Branco, the method successfully identified intriguing future scenarios, bolstering its applicability and value in real-world decision-making. It was also able to suggest interventions to reduce the number of vaccum breaks during production. 

Future research could focus on further refining our approach by exploring additional optimization strategies, such as meta-heuristic algorithms, to enhance its scalability and efficiency with larger datasets. Sometimes, the genetic algorithm converges to local minima and cannot evolve the population further. Other optimization techniques or alternatives to handle early convergence could be explored. Additionally, investigating the potential integration of domain-specific knowledge could improve accuracy and interpretability. 

\section*{Code and Data availability}
The code and data used for the machine-learning analyses, available for non-commercial use, has been deposited at \linebreak \url{https://doi.org/10.6084/m9.figshare.27609555}~\cite{zuin_2024}.

\section*{Acknowledgements}
This work was funded by the authors' individual grants from Kunumi. M. Dias Branco kindly granted access to the database needed for the development of the techniques in this study.

\bibliographystyle{IEEEtran}  
\bibliography{bibfile} 

\end{document}